# Inclusive Review on Advances in Masked Human Face Recognition Technologies

**Ali Haitham Abdul Amir [1], Zainab N. Nemer [2]**
[1,2] Department of Computer Science, Basra University, Basra, Iraq

**ABSTRACT**

Masked Face Recognition (MFR) is an increasingly important area in biometric recognition technologies, especially with the widespread use of masks as a result of the COVID-19 pandemic. This development has created new challenges for facial recognition systems due to the partial concealment of basic facial features. This paper aims to provide a comprehensive review of the latest developments in the field, with a focus on deep learning techniques, especially convolutional neural networks (CNNs) and twin networks (Siamese networks), which have played a pivotal role in improving the accuracy of covering face recognition. The paper discusses the most prominent challenges, which include changes in lighting, different facial positions, partial concealment, and the impact of mask types on the performance of systems. It also reviews advanced technologies developed to overcome these challenges, including data enhancement using artificial databases and multimedia methods to improve the ability of systems to generalize. In addition, the paper highlights advance in deep network design, feature extraction techniques, evaluation criteria, and data sets used in this area. Moreover, it reviews the various applications of masked face recognition in the fields of security and medicine, highlighting the growing importance of these systems in light of recurrent health crises and increasing security threats. Finally, the paper focuses on future research trends such as developing more efficient algorithms and integrating multimedia technologies to improve the performance of recognition systems in real-world environments and expand their applications.

***Corresponding Author:***

**Ali Haitham Abdul Amir**
**Zainab N. Nemer**
Department of Computer Science, Basra University, Basra, Iraq
Email: pgs.ali.hatham@uobasrah.edu.iq
Email: zainab.nemer@uobasrah.edu.iq

## 1. INTRODUCTION

It relies on recognition systems to identify individuals for various purposes. Examples of such systems include biometric, voice, graphic, letter, image, and facial recognition. Security systems primarily utilize biometric systems for access control [1]. Typically, face recognition systems identify basic facial features like the mouth, nose, and eyes, focusing on unobscured faces. However, certain cases and events necessitate individuals to wear masks that either partially obscure or completely contour their faces. These common scenarios include epidemics, laboratories, medical operations, and severe contamination. According to the Centers for Disease Control and Prevention and the World Health Organization, wearing masks and maintaining social distancing are the best ways to protect people from COVID-19 and prevent its spread [2]. As a result, most countries are making protective masks mandatory in public, which calls for examining the effectiveness of facial recognition systems in the presence of masked faces. Current facial recognition systems are ill-equipped to handle masked faces, as masks obstruct crucial facial features like the mouth and nose, making it difficult to accurately identify people. This poses a significant challenge in the field, highlighting the need for the development of techniques and algorithms capable of handling these situations. Potential solutions include the use of new systems based on residual facial features or deep learning techniques that can handle the differences caused by masks [3].

Face recognition, the automatic identification of individuals from images, is widely used in various applications. Many everyday uses, such as passport checks, smart doors, access control, voter identity verification, and criminal investigations, rely on face recognition for accurate and automated documentation. It has gained significant attention as a distinctive and reliable biometric technology, surpassing other methods like passwords and fingerprints. Governments around the world are also showing great interest in facial recognition systems to improve security in public places like parks, airports, bus stations, and train stations. Despite facing challenges in the past, such as varying facial illumination, diverse expressions, and angle differences, significant advances in deep learning have made face recognition more accurate and efficient in real-world applications. Over recent years, facial recognition technology has seen substantial progress, becoming one of the most studied topics in the field. Rapid advancements in machine learning have significantly reduced the challenges associated with face recognition. Deep learning has achieved remarkable progress in computer vision, including in areas like object recognition, classification, segmentation, and especially face recognition and verification [4][5].

The use of masked face recognition (MFR) greatly increases the complexity of the recognition process for people who wear face masks. This process becomes more difficult due to partial blocking and differences in appearance caused by facial masking. Masks hide important features such as the chin, lips, and nose, and the wide variety of styles, sizes, and colors makes identification more difficult. New strategies that show signs of success in solving MFR problems are deep learning techniques. Even when masks partially obscure facial features, we can train algorithms for face recognition. Researchers have proposed several deep learning-based procedures for masked face recognition, such as holistic approaches, mask-based approaches, and mask exclusion methods. Holistic methods in deep learning models use attention techniques to recognize complete facial features. Mask exclusion methods allow models to recognize the head, eyes, and some other features of a face that is covered by a mask. Techniques such as convolutional neural networks (CNN), Siamese neural networks (SNN), and YOLO [6] demonstrate the use of mask-based systems to generate adversarial generative networks (GANs), which in turn generate realistic faces in color images, thereby facilitating later recognition. However, despite the improvements made by deep learning techniques over many of the available public datasets, there is still a significant gap in the application of these models in real-world scenarios. Despite improved performance in controlled environments, challenges such as dealing with new mask types or changing environmental conditions still require more nuanced solutions before these models can be widely adopted in the real world [7].

## 2. Related work

This section examines earlier research in the domains of face mask recognition (FMR) and mask face recognition (MFR). In order to place the present research within the body of previous literature, this paragraph aims to provide a concise summary of several surveys in this field.

Different studies in the field of facial recognition address various techniques to tackle the challenges facing these systems, such as partial facial camouflage, one sample per person, and facial expression changes. For example, the study by Lahasan Bader et al. addresses these fundamental challenges and analyzes different techniques to overcome them, helping to improve facial recognition in real-world conditions. Liu and others proposed a way to improve feature representation by learning features firsthand, enhancing the system's ability to distinguish between individuals. On the other hand, in the SphereFace method, Win et al. used spherical spaces to improve feature representation, while Ding et al. in ArcFace focused on reducing the angular margin between categories to increase discrimination effectiveness. Similarly, Lin, Zhao, and others introduced a facial recognition system using deep learning with quantization techniques, extracting features using a convolutional neural network and then quantizing them using the "Bag-of-Features" framework. These studies represent important advances in facial recognition, as each contributes to addressing specific challenges and enhancing performance in practical applications, paving the way for the development of more integrated technologies in the future.

Hariri [2] conducted a study to improve the recognition of masked faces using deep learning techniques. In this study, occlusion removal, which begins with removing mask-covered areas of the face, was used to deal with the challenge of having masks on certain parts of the face, such as the mouth or nose. Next, we extracted deep features using convolutional neural networks (CNNs) such as VGG-16, AlexNet, and ResNet-50. Large datasets pre-train these networks, enabling them to extract hierarchical properties from images effectively. We used multilayered neural networks (MLP) for classification after extracting the features, which helped correlate the extracted properties with the identified identity. We tested the method on the real-world masked faces database (RMFRD) and found that ResNet-50 had the highest facial recognition rate compared to other models like VGG-16 and AlexNet. This suggests that the ResNet-50 architecture proved to be more effective in handling mask-induced occlusion. However, it is important to consider some critical points that may improve this research. Firstly, examine the implementation of occlusion removal and identify any potential effects or disruptions that could impact the recognition accuracy following this technique. A comparison between VGG-16, AlexNet, and ResNet-50 provides an idea of how different models perform in the task of recognizing masked faces, but it's helpful to discuss why ResNet-50 outperforms other models, such as mesh depth or residual connections. Furthermore, while the use of MLP for classification is a common technique, more sophisticated methods such as transformer-based networks or attention mechanisms can be considered to improve the results. Finally, while the real-world Masked Faces Database (RMFRD) provides realistic data, it's important to consider factors such as variety, lighting conditions, and types of masks used, as these could potentially impact the results. Overall, Hariri's approach shows promising results, but it is useful to evaluate the techniques used more deeply and consider areas of improvement that may be useful in future research.

The study by Sony et al. [10] focused on developing a model to detect in real-time whether a person is wearing a helmet or not, thus identifying irregularities. We implemented the project using TensorFlow, Keras, and OpenCV. The model showed significant improvements over previous systems, especially when it faced situations where people wore face coverings such as masks or face shields—a problem that was causing prediction errors in previous models. TensorFlow and Keras integration demonstrates the model's reliance on deep learning for helmet processing, while OpenCV handles image processing. Although the model has demonstrated improvements in handling face coverings, it is crucial to evaluate its performance in a variety of environmental conditions, including different lighting, angles, and obscurations. We can compare this model with other helmet detection systems in terms of accuracy, speed, and adaptability to various conditions, particularly when individuals wear face coverings that could impede the recognition process. Other models demonstrate a significant advancement in helmet detection technology by addressing the issue of face coverings, a prevalent issue in many computer vision-based models. However, the model requires further analysis and testing in various conditions to confirm its effectiveness in practical applications.

Consequently, researchers have focused on leveraging the extensive proliferation of deep learning and the commendable efficacy of its models. They retrieved features for masked face recognition, and Zhang et al. developed a training technique based on a dual-branch approach to direct the network's attention to the unobstructed areas of the face. We jointly optimized the two branches to improve feature extraction in non-occluded areas. We proposed a face recognition network to identify impacted faces obscured by occlusion and heavy noise. We extracted the noise-susceptible NRLBP, FLBP, and LBP characteristics. Subsequently, the mean of the three characteristics served as the input for the proposed CNN classifier. The proposed complete face recovery GAN restored occluded areas and collapsed textures by leveraging the textural and structural disparities between two images, eliminating the need for human intervention or paired data to establish the ground truth through the selection of occlusion-free images [11].

The study of Alzu'bi Ahmad et al. [12] comprehensively evaluated developments in masked face recognition (MFR) research, which has undergone significant development in recent years. The paper systematically examined different approaches and strategies in the field of MFR, with a particular focus on deep learning techniques. The study aims to provide valuable insights into the progress of MFR systems through a comprehensive review of current work, and also addressed the main data sets used in research and the evaluation criteria used to measure the effectiveness of these systems. The study also provides a complete framework for evaluating different methodologies while highlighting current challenges and future research opportunities in the field.

## 3. The mask face recognition sequence

This section illustrates the usual development of MFR systems via a series of intricate stages, as seen in Figure 1. The general technique primarily relies on models of deep learning that are extensively used to extract distinguishing aspects of masked faces. This pipeline illustrates the essential phases involved in constructing the ultimate method of identification, as elaborated on the subsequent subdivisions.

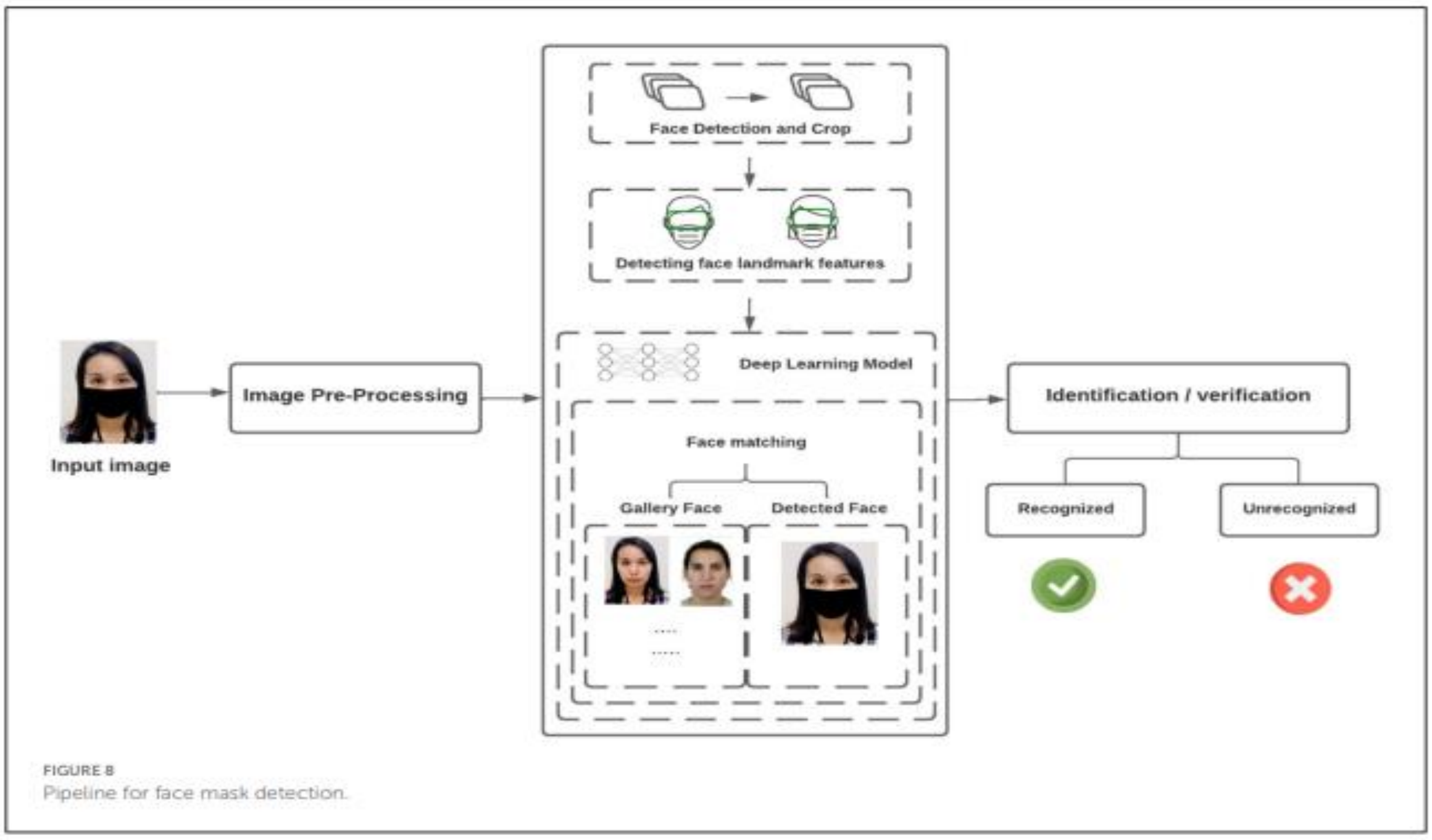

Figure 1. The foundation for masked face recognition shown graphically [13].

Initially, a compilation of original masked photos and their associated ground-truth photographs is assembled. This often involves categorising them into distinct directories for model training, validation, and testing. This is followed by preprocessing processes including data augmentation and picture segmentation. A set of essential facial traits is retrieved via one or more deep learning models, often pretrained on general-purpose photos and then fine-tuned on a fresh dataset, specifically masked faces. These traits must be sufficiently discriminative to reliably identify face masks. A face unmasking process is then used to recover the concealed face and provide an assessment of the original visage. The anticipated face is compared against the original ground-truth faces to see whether a certain individual is recognised or confirmed.

### 3.1. Deep Learning

The phrases artificial intelligence, machine learning (ML), deep learning (DL), and neural networks are intricately interconnected. Deep learning is a subset of machine learning, both of which are included under artificial intelligence; the distinctions between them will be elucidated in the subsequent talks and analysis section. The link between deep learning and neural networks is such that deep learning is defined when a neural network has more than three layers, especially including at least one hidden layer in addition to the fundamental input and output layers, which need the application of a series of learning algorithms. In summary, deep learning depends on a neural network characterised by several hidden layers, also referred to as deep neural networks [14].

Numerous established techniques have been suggested to identify human faces using handcrafted local or global characteristics, including LBP, SIFT, and Gabor [15], [16], [17]. Nonetheless, these holistic techniques are hindered by their inability to manage face alterations that diverge from their original assumptions. Subsequently, shallow image representations, such as learning-based dictionary descriptors, were added to enhance the efficacy and compactness of prior methodologies [18]. Despite advancements in accuracy, these superficial representations continue to exhibit poor resilience in real-world applications and instability in response to alterations in face appearance. Post-2010, deep learning techniques were swiftly advanced and used via several deep layers to extract characteristics and modify images. They demonstrated superiority in acquiring various tiers of facial representations corresponding to varying degrees of abstraction, exhibiting a robust consistency in face alterations, including lighting, articulation, posture, and disguise. Deep learning models may integrate low-level and high-level abstractions to effectively represent and recognise a consistent visual identity with high precision. This section introduces prevalent deep learning algorithms used for masked face recognition [19].

The use of deep learning in facial recognition will improve accuracy, elevate performance, and provide significant outcomes due to its capacity to categorise a vast array of unidentified faces. Likewise, the deep learning methodology enhanced certain traditional techniques using CNNs taught via a supervised way [20]. Diverse methodologies have used CNN for facial recognition, categorised by prominent architectural types like AlexNet, VGGNet, GoogleNet, LeNet, and ResNet. The amalgamation of face recognition with deep learning technologies has yielded several advantages, including enhanced security in commercial operations and public areas, prompting the majority of researchers to focus their efforts in this domain. This section integrates two essential domains that allow the effective and precise detection and recognition of masked faces via feature extraction and convolutional neural networks (CNNs). Additionally, other datasets, including real-world masked faces, may be used to train the model for face recognition. Figure 2 illustrates three basic steps that provide persuasive facial recognition using deep learning.

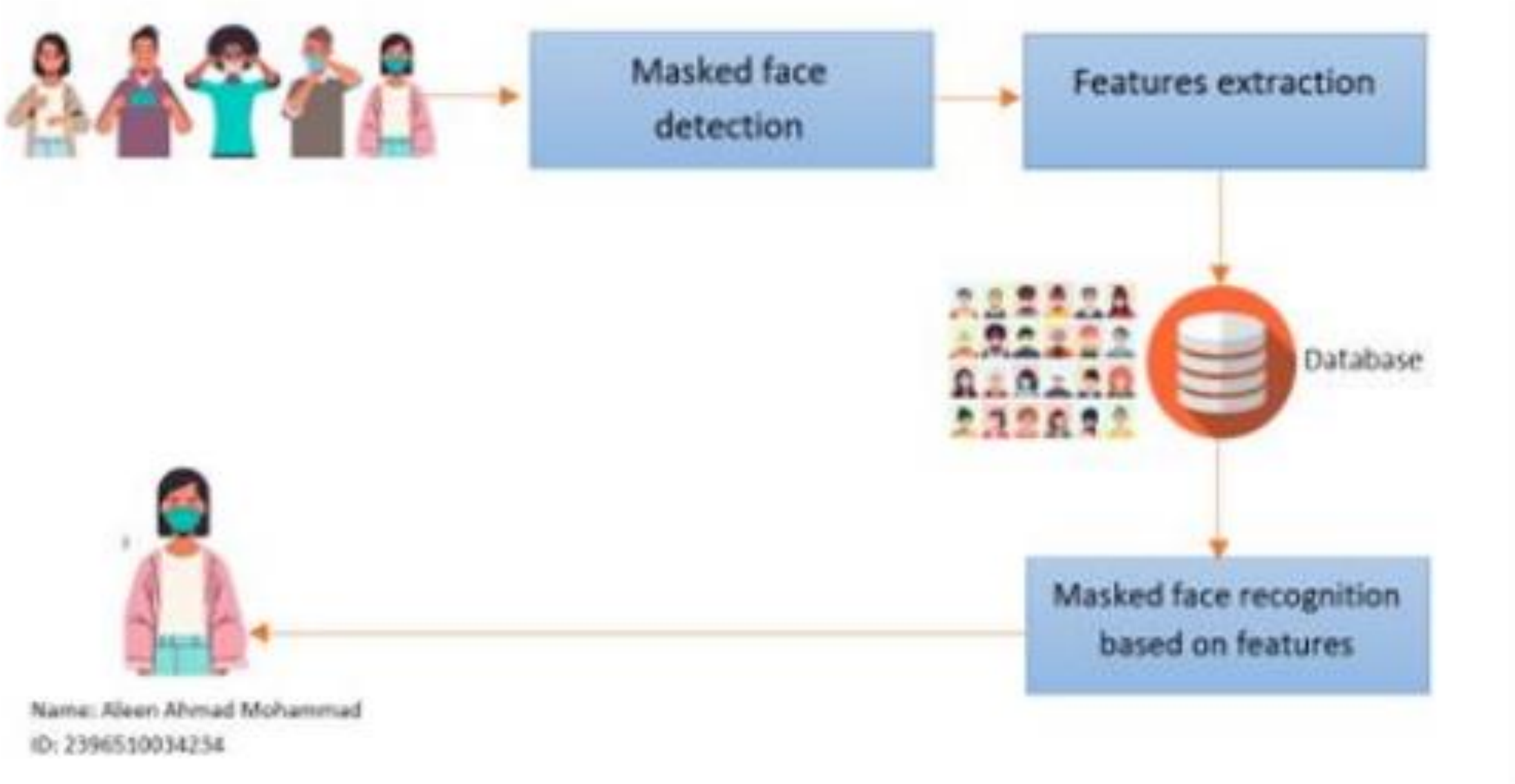

Figure 2. Three main processes that create masked face recognition based on deep learning[20]

### 3.1.1. Convolutional Neural Networks

In several domains, such as picture categorization, recognition, retrieval, and object detection, convolutional neural networks (CNNs) have shown to be the most successful neural networks. Input, convolutional, subsampling, fully linked, and output layers typically make up a Convolutional Neural Network (CNN), with the purpose of controlling shift, scale, and distortion. Lighting, posture, facial expression, and age are just a few of the many intra-class changes that they are able to successfully detect using training data. We have trained CNN-based models on many large-scale face datasets, and we have deployed them widely [21] [22]

One of the pretrained architectures most often used in face recognition applications is AlexNet [23]. Even with large datasets, AlexNet was able to minimize errors and shorten training times by using integrated graphics processing units (GPUs).

Several CNN-based systems, such as VGG16 and VGG19 [24] and the Face-mask Detection System [25], are widely utilized in computer vision applications, such as face recognition. Convolutional representations or features are often offered by VGG-based models. Despite the significant accuracy achieved, they are limited by the complexity and length of training. The usage of deeper neural networks became necessary as photo recognition became more sophisticated over time. However, adding more layers to the networks makes training more difficult and may sometimes lead to a drop in accuracy.

The remaining network (ResNet), which adds more layers to improve performance and accuracy, was developed to overcome this challenge for both the hidden face [26] and the displayed [26]. Although new layers can have complex characteristics, their addition must be evaluated experimentally to counteract any decline in model performance. MobileNet [27], a lightweight and lightweight deep neural network with a simplified configuration, is often used in face recognition applications. Done Accelerated model calculations, and the structure showed excellent performance with super parameters. CNN-based eminents such as Inception and their forms are innovative in that they use modules or blocks to build networks using convolutional layers rather than just stacking them. An improved version of Inception called Xception [28] uses detachable convolutions by depth [21] instead of its units. Figure 3 shows the structure of CNN Face detection.

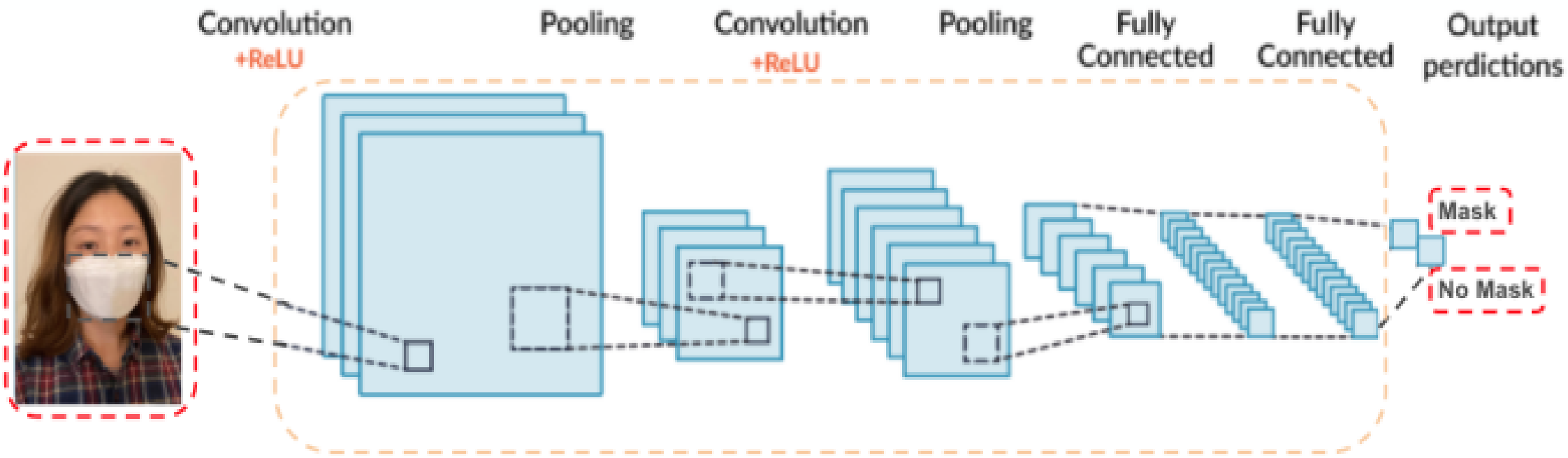

Figure 3. CNN architecture for mask detection [29]

### 3.1.2. You Only Live Once (YOLO)

YOLO is a real-time object detection system recognized for its rapidity and precision, making it suitable for applications such as face mask identification. S. Singh used YOLOv3 and Faster R-CNN on a dataset including 7,500 pictures, whilst X. Jiang enhanced YOLOv3's efficacy by using SE blocks, GIoU Loss, and Focal Loss. J Yu improved YOLOv4 by integrating CSPDarkNet53 and PANet, attaining an accuracy of 98.3%. J Ieamsaard executed YOLOv5, achieving 96.5% accuracy after 300 epochs, while TN Pham presented the YOLOv5s-CA and C3CA models for enhanced performance. P Wu introduced the FMDYolo framework with Res2Net-101 for improved efficiency, while H Zhang created AI-Yolo by integrating SK units, SPP, and CIoU to augment accuracy. Finally, S Tamang illustrated YOLOv8's advantage compared to YOLOv5 for mask recognition, making it optimal for real-time applications [30].

The YOLO technology employs a single neural network to evaluate the whole picture instead of using fragmented methods. Segments the picture into grids and delineates the bounding boxes around the items, along with classification probabilities. YOLO is an advanced technique for real-time object detection. In masked face detection, YOLO-v5 is used to categorize individuals as mask wearers or non-wearers and to ascertain if the mask adequately covers the nose and mouth. Research has shown the efficacy of YOLO-v5 in this endeavor, obtaining an accuracy of 90.37% using the YOLO-v5s model. YOLO-v5 is efficient and adept at processing faces of varying dimensions, however it has difficulties with little items [31].

The YOLO approach is rapid and reliable but struggles to detect small objects and differentiate them from two-flight explorers. The network started with iterations such as YOLOv2, which had enhancements such batch normalization and high-definition workbooks, followed by YOLOv3, which used Darknet-53 and digital squares. I endorsed YOLOv8, which exhibits improvements in precision and efficacy. YOLO, SSD, and RetinaNet networks were promoted with several layers and various feature map dimensions[32] .
Ahmed and colleagues relied on the YOLOv3 model to detect people in high resolution in videos. Suryadi et al. also used three classification models to track social distancing: YOLOv3, a watered-down version of it, and MobileNetSSD, which combines a multi-box detector (SSD) and MobileNet. These models were tested on videos from surveillance cameras at the Oxford site using the MS COCO dataset, where YOLOv provedIt outperforms the lightweight version and MobileNetSSD. On the other hand, Rezaei and Azrami developed the DeepSOCIAL model, which relies on YOLOv4 with CSPDarknet53, to improve the speed and accuracy of detecting people and estimating distances between them [13]. Figure 4 shows the structure of YOLO.

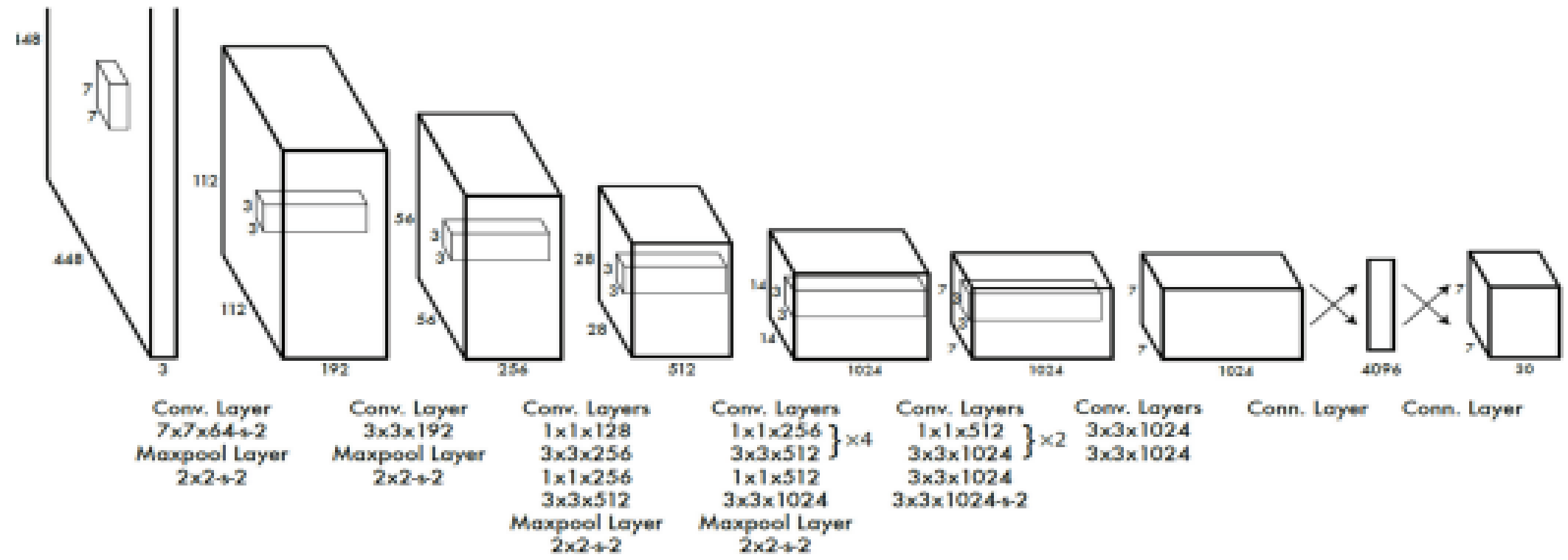

Figure 4. YOLO architecture[32].

### 3.1.3. Deep Belief Network Deep

A deep belief network (DBN) has numerous layers of interconnected hidden units, with no connections across units within the same layer. It generally comprises a sequence of restricted Boltzmann machines (RBMs) or autoencoders, whereby each hidden sub-layer functions as a visible layer for the subsequent hidden sub-layer, culminating in a softmax layer used for classification. Deep Belief Networks (DBNs) have been used in the fields of Face Recognition [33], Facial Expression Recognition[34], and Occluded Face Recognition [30]. The Boosted Deep Belief Network (BDBN) aims to consolidate the phases of feature learning, selection, and workbook building into a single iterative process. BDBN analyzes and identifies characteristics associated with changes in form and appearance of facial expressions, using them to enhance the workbook. The workbook's performance and the chosen features are progressively enhanced via a standardized fine-tuning approach. Research using public datasets has shown substantial advancements in the understanding of facial expressions [34] . Exact regulation between the material and the cutting instrument is crucial for attaining precise outcomes in manufacturing operations. Any change in cutting parameters, including speed, feed, or cutting depth, results in tool deterioration and unfinished work. The Deep Belief Network (DBN) was created to identify six instrument states (one intact and five damaged) by analyzing image-based vibration signals in real time for the purpose of diagnosing tool failure. The model is developed and evaluated with diverse datasets [35].

### 3.1.4. Generative Adversarial Networks

Without extensively annotated training datasets, generative adversarial networks (GANs) may infer and incorporate the underlying patterns from incoming data on their own. A GAN is made up of two neural networks: one for generation and another for discrimination. In order to produce new traits, the generator uses stochastic data, which consists of random values drawn from a certain distribution. In its role as a binary classifier, the discriminator checks the generated features for signs of phoney or realness. The adversarial training framework of GANs is the reason for its adversarial name. In this context, the generator and discriminator play a minimax game to optimize competing loss functions. Facial synthesis [36], makeup-invariant face recognition [37], masked face recognition, age-separated faces [38], pose-invariant face recognition [39], face manipulation [40], and cross-age face recognition [41] are just a few of the common face recognition problems that Generative Adversarial Networks (GANs) have successfully addressed.

## 3.2 machine learning

Algorithms for Machine Learning Classification algorithms are used to categorise things of diverse sorts. They assist in categorising items into analogous or disparate groupings. These algorithms are essential to face recognition. They assist in classifying the pictures and assessing their interrelations. Our exploratory research employs six distinct conventional machine learning classification techniques for experimentation [42].

- **Support Vector Classifier (SVC)**: Assistance the Vector Classifier facilitates binary classification tasks and may be used for multi-class challenges. SVC preserves robust generalisation by non-linearly mapping inputs to high-dimensional feature spaces and establishing linear decision boundaries.
- **Linear Discriminant Analysis (LDA):** Analysis of Linear Discriminants PCA, or principal component analysis and other classification approaches are well recognised. As indicated by its nomenclature, LDA functions as a linear classifier. LDA is a very effective technique for dimensionality reduction. It is often used to extract features in pattern classification tasks.
- **K-Nearest Neighbors (KNN)**: It is used mostly for addressing data mining and picture classification challenges. KNN functions as both a classifier and a regressor; however, this research utilises it only as a classifier.
- **Decision Trees (DT)**: Decision Trees embody a flowchart-like configuration. Decision Trees differ from Support Vector Classifiers and neural networks in that they do not impose statistical assumptions on the inputs or the size of the data.
- **Logistic Regression (LR)**: Logistic Regression facilitates the modelling of the likelihood of a certain class or existing classes. Notwithstanding its designation, it functions as a classifier rather than a regressor. This is a simple and very effective approach used for binary and linear classification tasks. It is the most used machine learning model in the business.
- **Naïve Bayes (NB)**: is effective for both binary and multiclass classification tasks. It is expected to excel with categorical inputs relative to numerical variables.

### 3.3. Feature extraction

An essential part of face recognition is feature extraction, which seeks to isolate and identify unique facial characteristics including texture, eyes, mouth, and nose. Masks and other facial impediments make this procedure more difficult and need adjustments to current face recognition systems to guarantee accurate and strong feature representation. There are two main schools of thought when it comes to feature extraction methods used for masked face recognition: the deep representation school and the shallow representation school. Accurate identification is made possible by capturing distinctive patterns from face photos using various approaches. A lot of people still utilize old-school techniques like Haar Cascades and Local Binary Patterns (LBP), however they can't manage the variety and complexity of modern picture data. New deep learning neural network models, such as ResNet and EfficientNet, use their complex architectures to examine fine features in pictures, greatly enhancing performance in cases when faces are partly obscured.[43]

Several techniques have been explored and evaluated for facial feature extraction using deep learning approaches. Li et al. [44] highlighted that masked faces often contain region-specific information related to the mask, necessitating tailored modeling strategies. They proposed the creation of two distinct centers for each class: one for full-face images and another for masked face images. Song et al. [12] introduced a multi-stage mask learning approach that leverages CNNs to identify and exclude compromised features from recognition processes. Various attention-aware and context-aware methods have employed auxiliary sub-networks to focus on critical facial regions for enhanced feature extraction. Moreover, deep graph convolutional networks (GCNs) have been applied to tasks such as masked face detection, reconstruction, and recognition, utilizing graph-based representations of facial images. GCNs have demonstrated strong capabilities in analyzing facial features using spatial or spectral filters adapted to uniform or static graph structures. However, the effectiveness of these networks can be constrained by the depth of GCN layers and the computational complexity involved in processing.

Siamese Networks are an optimal solution for face recognition problems, simultaneously analysing the attributes of two photos to ascertain if they correspond to the same individual. These networks are trained on a dataset of photos of masked faces, enabling them to comprehend the subtleties of facial characteristics. Generative neural networks use GANs (Generative Adversarial Networks) to enhance the data employed in model training. For instance, GANs may create pictures of realistic faces, augmenting data variety and improving the model's capacity to properly identify masked faces [45] The characteristics of 3D space have also been examined for the purpose of occluded or masked 3D face recognition. 3D face recognition techniques emulate human visual perception and comprehension of facial traits, hence enhancing the efficacy of current 2D recognition systems. The three-dimensional facial characteristics are resilient to many alterations, including changes in light, facial emotions, and head orientations.

### 3.4. **Data augmentation**

Data augmentation is a prevalent method to optimizing the use of an informational repository. To augment the training set, little alterations to the photos are implemented, including modest changes made to the supplied photos, including scaling, rotating, and translating. Data augmentation enhances the efficacy and outcomes of machine learning models by generating novel and diverse instances for training datasets. Machine learning algorithms necessitate substantial datasets for effective training, and the scarcity of available datasets, particularly in emerging research domains, necessitates data augmentation to enhance the diversity of existing datasets through modifications. Some methods used to increase data diversity:

- **Cutout and Random Erasing**: These procedures include occluding certain areas of the face, including the nose and mouth, in training pictures. This replicates the impact of masks, aiding the algorithm in learning to identify faces with partial obstructions. Studies demonstrate that using these techniques may enhance the resilience of masked face recognition systems [46]
- **Synthetic Mask Generation**: This methodology generates synthetic masks using generative models or graphical techniques for application to existing facial photos. This facilitates the equilibrium of datasets by guaranteeing sufficient representation of both masked and unmasked faces, hence augmenting the model's generalisation capability [47].
- **Blur Augmentation**: Techniques like Gaussian blur are used to replicate the impact of masks on facial clarity. This enhancement prompts the model to concentrate on unique characteristics that persist even in the presence of a mask [48].
- **Geometric Transformations**: This entails the application of elastic distortions, random flips, and rotations to pictures, training the model to identify masked faces from many angles and viewpoints.

This approach is especially beneficial in practical situations where facial orientations may differ significantly [49].

- **Random Noise**: This approach is incorporating random noise into photos to replicate diverse lighting situations or catch inaccuracies. It assists the model in identifying obscured faces across various contexts [5].
- **Color Jittering**: This enhancement alters the brightness, contrast, or saturation of the photographs. It emulates the impact of varying illumination conditions and augments dataset variety, hence improving the model's resilience [50]
- **Random Cropping**: This method randomly extracts segments of the picture, aiding the model in identifying masked faces despite obstructions or ambiguities in facial features [51].
- **Random Rotation**: This technique entails rotating pictures by a certain angle, aiding the model in processing faces from diverse orientations, which is prevalent in practical applications [52].

### 3.5 Image preprocessing

One of the quickly advancing technologies nowadays is a technique that involves many operations on the input picture to get informative insights from improved images. Input pictures are represented in pixels, with each pixel corresponding to the three colors red, green, and blue (RGB), and sometimes black and white. It is a kind of signal processing and encompasses two methodologies: two types of picture processing: digital and analogue. Physical copies, such as prints and pictures, rely on analog image processing. Processing digital images involves three critical stages: processing, augmentation, and presentation, as well as knowledge extraction [53].

Image pre-processing enhances the efficiency of the trained models in the dataset by serving as a method of data cleansing. Due to significant variations in contrast and exposure in raw images taken in real life, image pre-processing is necessary to guarantee the accuracy of facial recognition and identification of face-mask-wearing status [54]. Only a limited number of datasets are publicly accessible. Consequently, it is essential to include synthetic images including diverse face masks into the testbed and enhance the generalization skills of models for deep learning. The purpose of the data pre-processing procedures is to clean up the available data and eliminate items that may impede performance, such as smoke and pictures that are distorted, fuzzy, or lack critical components.

### 3.6. Face recognition

In the field of image analysis, face recognition is one of the most effective uses [55]. As computers and computing systems become increasingly efficient and affordable, humans are able to recognize faces easily. As a result, digital image processing has gained a lot of attention in many computer applications and human-computer interaction [56]. It is used to include face information in electronic passports worldwide. In addition, it serves as a user interface Normal in consumer electronics, entertainment, security and law enforcement. By identifying the user and providing customized services for consumer devices, it improves the user experience. Every processing step in the face recognition system is performed to meet the application requirements for optimal performance. Face recognition is the process of identifying a person in an input image by comparing it to a database of saved faces. Face recognition is influenced by a number of factors, including size, shape, position, blockage, and lighting. It can be used for simple and complex purposes. In order to create a unique digital code, master face recognition looks for distinctive signs including width along the nose, widening of the eyes, angle and depth of the jaw, vertical dimension of the cheekbones, and distance between the eyes. Then, the algorithm uses these numeric codes to compare the two images and see how similar they are. The image source for face recognition includes both live video feeds from cameras and archived images from various sources. Parts of a facial recognition system include the ability to recognize faces, extract features from those faces, and then recognize those faces [55]. Figure 5 shows the process of face recognition.

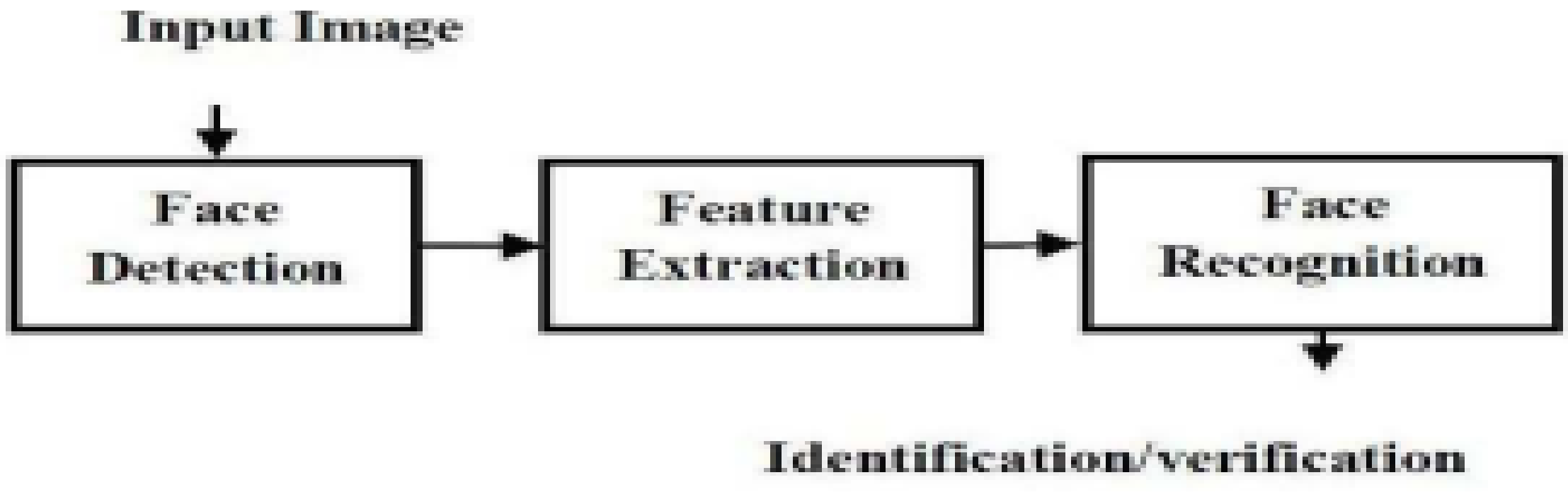

Figure 5. Processes in face recognition system [57]

### 3.7. Masked face recognition

In masked face recognition, the goal is to get the system to identify a person even while their face is obscured. Because masks obscure over half of the face and it's hard to tell someone's identity from features like the eyes, eyebrows, and forehead alone, masked face identification is a challenging challenge. As a result, training the optimal model to correctly identify a person using masked facial recognition often requires massive computations and datasets [58].

As a safe and effective way to verify someone's identification while they're wearing a mask, MFR shows great promise in several fields. Innovative solutions to modern problems may be realised from this potential. To demonstrate MFR's versatility and importance, this section delves into its many potential uses. Institutions of Higher Learning, Retail, Customer Service, Human-Computer Interaction, and Workplace Security and Access Control [8]

## 4. Datasets

The widely used standard datasets for a variety of masked face tasks, including as face mask identification, mask face recognition, and face unmasking, are examined in this section. to guarantee a thorough analysis and in-depth discussion of some of the most often used datasets in face mask detection methods, taking into account their size and content.

One of the most extensive publicly available datasets for masked face recognition (MFR) is the Real-World Masked Face Recognition Dataset [59]. With 90,000 unmasked faces and 5,000 masked faces from 525 individuals, RMFRD offers a substantial resource for MFR model training and evaluation. Researchers created SMFRD [60] to increase the dataset's diversity, which included 500,000 images of 10,000 people's faces that were artificially disguised and taken from the Internet. There are two parts to the Masked Face Segmentation and Recognition (MFSR) dataset. 9,742 images of masked faces from the Internet are included in the first part, together with meticulously labeled masked area segmentation comments. Of the 11,615 photos in the second section, which depict 1,004 individuals, 704 were taken from actual collections; the other photos were downloaded from the Internet. At least one picture of both masked and unmasked faces is included in each identification. IMDb is the source of the Webface dataset [61], which consists of 500,000 images that correspond to 10,000 identities. 16,128 images representing 28 identities in nine poses and 64 lighting situations make up the Extend Yela B collection. An improved version of CASIA-FaceV5[62], CASIA-FaceV5_m [62], has 2500 images of 500 Asian people, five of whom are featured in each image. 9,205 images from the Properly Wearing Masked Face Detection Dataset (PWMFD) [63] have been categorized into three groups. Furthermore, TFM's vast private collection of 107,598 images contrasts with databases like Moxa3K, which have 3,000 photos. 7,804 images make up the Wearing Mask Detection (WMD) dataset, which is used to train detection algorithms. The 38,145 photographs in the Wearing Mask Classification (WMC) dataset are divided into two categories: 19,590 pictures of masked faces and 18,555 background samples. There are 4,000 images with 126 different IDs in the AR collection. The 6,264 images in the Bias-Aware Face Mask Detection (BAFMD)[64] collection include 3,118 uncovered faces and 13,492 masked faces. The fact that there are several faces in each image is important. Three thousand images of 300 people were taken from the Internet and included in the Masked Faces in the Wild (MFW) small dataset [65]. Each person had five shots of their masked faces and five of their unmasked faces. There are 45 participants in the Masked Face Database (MFD), which has 990 images of both men and women. There are over 50,000 images in the Labelled Faces in the Wild (LFW) collection. For training, Golwalkar et al. [66] employed 204 images and the masked faces of 13 people. They utilized identical face images for the examination, including 25 shots of each person. Furthermore, in [67], the LFW-SM variant dataset was made public. It included 13,233 images of 5,749 people and supplemented the LFW dataset with simulated masks. The VGGFace2 [68] dataset, which included 3 million photographs of 9,131 people, with around 362 images per person, was utilized as training material for a number of MFR algorithms.

**Table 1: Summary of the mask face recognition datasets.**

| dataset | Size | Identities |
|---|---|---|
| RMFRD | 95,000 | 525 |
| SMFRD | 500,000 | 10,000 |
| Extend Yela B | 16,128 | 28 |
| MFSR | 11,615 | 1004 |
| MFR2 | 269 | 53 |
| VGG-Face2_m | 666,800 | 8335 |
| COMASK20 | 2754 | 300 |
| MFW | 6006 | 2 |
| BAFMD | 6264 | 2 |
| LFW-SM | 13,233 | 5749 |
| WMC | 38,145 | 2 |
| WMD | 7804 | 1 |

## 5. Evaluation Metrics

The traditional evaluation tools used in MFR, FMR, and face unmasking will be detailed in this section. To determine the effectiveness of models in real-world applications, it is essential to evaluate their performance in these domains. As a result, accuracy, robustness, and efficiency are evaluated using a variety of evaluation criteria and benchmarking approaches. This presentation will take a look at the main indicators and approaches to benchmarking that these positions utilize for evaluation.

**1.Accuracy** is a fundamental evaluative metric used in several fields, including facial recognition applications. It signifies the proportion of correct predictions to the total number of samples, formally defined as shown in Equation (1) [69].

$$\text{Accuracy} = \frac{TP + TN}{TP + TN + FP + FN} \quad (1)$$

2. **Precision** is quantifying the ratio of correct positive identifications to the total positive matches found, formally represented as shown in Equation (2) [70]

$$\text{Precision} = \frac{TP}{TP + FP} \quad (2)$$

**3. Recall** include the measurement of the ratio of correctly identified positive matches to all positive matches found, as formally stated by Equation (2) [70].

$$\text{Recall} = \frac{TP}{TP + FN} \quad (3)$$

4. **F1-Score** the symphony of recall and accuracy, is an essential evaluation metric. All aspects of the face recognition model's performance, including false positives and false negatives, are evaluated by this metric. Because it provides a comprehensive assessment of model performance, the F1-Score is particularly useful for imbalanced datasets. The F1-Score is a more reliable indicator of a model's performance than accuracy, which could ignore certain types of mistakes. This is because it takes both types of mistakes into consideration. This calculation is shown in Equation (4). [69]

$$\text{F1} - \text{Score} = 2 \times \frac{\text{Precision} \times \text{Recall}}{\text{Precision} + \text{Recall}} \quad (4)$$

5. **ROC (Receiver Operating Characteristic)** ROC curves graphically show the trade-off between sensitivity (the rate of true positives) and specificity (the rate of true negatives) at various threshold levels. With the help of this graphic, we can find the sweet spot where the rates of true positives and false positives are equal. It is possible to assess a classification model's performance and make well-informed decisions on threshold selection by examining the ROC curve. [71].
6. **AUC (Area Under the Curve)** is an important evaluation metric for classification jobs as it gives a comprehensive review of how well a model performed. By comparing the model's performance at various threshold levels, we can see how well it can distinguish between positive and negative events. An increase in the area under the curve (AUC) shows that the model is more effective since it can discriminate better. The model's predictive power is as good as random chance when the area under the curve (AUC) is 0.5. AUC is widely utilized in performance evaluation across several domains and is crucial for determining how well categorization algorithms work.[71].
7. **Confusion Matrix**: provides a comprehensive analysis of the model's predictions, encompassing all of its outcomes, including accurate forecasts, incorrect predictions, and erroneous predictions. It may be utilized to determine various evaluation indicators and identify problem areas [70].
8. **FRR (False Rejection Rate)** indicates the likelihood of the system incorrectly rejecting a legitimate identity match, which is an important metric for evaluating the system's usability. A high False Rejection Rate (FRR) indicates lower usability due to the frequent denial of access to authorized users, hence the assessment is vital for determining the system's user-friendliness. If we want to make users happier and access protocols more efficient, we need to get the FRR down. As shown in Equation (5), the FRR may be calculated [72].

$$\text{FRR} = \frac{\text{FN}}{\text{FN} + \text{TP}} \tag{5}$$

9. **Specificity** The true negative rate is another name for this metric, which assesses how well the system can spot negative scenarios. Accurately identifying individuals who are not the intended targets is the unique capacity of the system that is tested. Applying Equation (6)[73] yields the mathematical determination of specificity. This metric improves the system's overall performance and reliability by shedding light on how well it classifies negatives.

$$\text{Speci f icity} = \frac{\text{TN}}{\text{TN} + \text{FP}} \tag{6}$$

10. **Geometric mean (G-Mean):** Finds the sweet spot where the majority and minority classes' categorization performance meet. Even when negative examples are correctly categorized, a low G-Mean shows that the model is not very good at categorizing positive ones. In order to avoid the negative class being overfit and the positive class being underfit, this method is crucial. An indicator of how well positive cases may be identified is the sensitivity. When evaluating the validity of negative scenarios, specificity is used in the other direction [74].

$$\text{Sensitivity} = \text{TP}/\ (\text{TP} + \text{FN}) \tag{8}$$

$$\text{Specificity} = \text{TN}/\ (\text{TN} + \text{FN}) \tag{9}$$

$$\text{G} - \text{Mean} = \sqrt{\text{Sensitivity} \times \text{Specificity}} \tag{10}$$

11. **Error rate (ERR):** The error rate or misclassification rate is the inverse of the accuracy measure. This measure quantifies the number of incorrectly identified samples from both the positive and negative classifications. It is susceptible to skewed data, analogous to the accuracy metric [75]. The calculation may be performed as follows:

$$\text{ERR} = 1 - \text{Accuracy} \tag{11}$$

12. **Equal error rate (EER):** To find out when the false rejection rate (FRR) is equal to the erroneous acceptance rate (FAR), one uses the Equal Error Rate (EER). The more accurate the system is, the lower the EER value. Equation (14) may be used to find the Equal Error Rate (EER), whereas Equations (12) and (13), respectively, can be used to get the False Acceptance Rate (FAR) and False Rejection Rate (FRR), as previously mentioned [76].

$$\text{FAR} = \frac{\text{FP}}{\text{FP} + \text{TN}} \tag{12}$$

$$\text{FRR} = \frac{\text{FN}}{\text{FN} + \text{TP}} \tag{13}$$

$$EER = \frac{FAR + FRR}{2} \quad (14)$$

13. **False discovery rate (FDR):** The expected proportion of positive classifications (rejections of the null hypothesis) to the total number of false-positive determinations (false discoveries). There are two types of positive results that add up to the total number of null hypothesis rejections [77]. You can figure out the FDR by doing this.

$$FDR = \frac{FP}{FP + TP} \quad (15)$$

14. **True positive rate (TPR):** The recall or true positive rate is the fraction of positive samples that were correctly recognized out of all the positive samples [78]. Here is how the computation may be done:

$$TPR = \frac{= TP}{FN + TP} \quad (16)$$

## 6. Challenges and Future Work

### 6.1. Challenges

Future research has several opportunities to enhance the existing capabilities of mask face recognition technology. Essential focal points should include the following We plan to contribute in the future via many endeavors, including:

1. In the context of hard-balanced data, the disparities in the number of teams across several categories are minimal, but insufficient to classify the data as entirely balanced or significantly imbalanced. In instances with a very under-represented class, the model may entirely exclude that class. The outcome is a model that functions inadequately.
2. Binary classification datasets for face mask detectors should include not only the detection of masks but also the manner in which they are worn. An improperly worn mask has equivalent health hazards to not wearing a mask at all. Consequently, categorising those who don incorrect face masks alongside those who wear acceptable masks is erroneous, and such individuals will persist in their detrimental practices. Furthermore, using these datasets in models to assess individuals' compliance with face mask use would provide inaccurate findings.
3. A crucial step in calibrating the model to identify masked faces is optimising parameters, whereby the most effective values for model parameters such as age count, learning rate, and batch size are determined to enhance model performance. The ideal efficacy of the learning algorithm is attained by the judicious selection of the most suitable hyperparameters, which will eventually influence the researcher's understanding of the algorithm's capabilities and limitations. Identifying the suitable hyperparameters is essential for effective training of the learning mechanism and attaining the desired performance metrics.
4. Because embossed masks are often sophisticated and include features like lips, noses, and chins, they are challenging for detection algorithms that are mainly intended to recognise surgical masks. As a result, it might be challenging to identify personal masks, which lowers accuracy and increases false positives.
5. Face mask detection serves as a preliminary procedure wherein the integration of mask detection and masked face recognition may improve the robustness of recognition systems. Treating face unmasking as a pre-processing step enables models to get enhanced accuracy in situations characterised by differing levels of face occlusion. This strategy will also assist in alleviating the difficulties associated with inconsistent face coverings.

### 6.2. Future Works

1. Future research might extend our suggested methodology to address other forms of obstructions, like sunglasses, scarves, and hats, which may similarly provide difficulties for face recognition systems.
2. Integrating additional biometric modalities, such as voice and fingerprint identification, may enhance the system's accuracy and dependability.
3. A further domain for future investigation is examining the privacy ramifications of masked facial recognition. With the widespread use of masks, apprehensions over the implications of face recognition technology on privacy have intensified.
4. Detection of social distance and an alert that goes off if a person isn't wearing a face mask correctly are two of our next projects.
5. Enhance the suggested system to include the identification of masked faces to augment security and reliability. Development of the suggested technology for use in electronic security portals

## 7. CONCLUSION

This paper offers a thorough analysis of current developments in deep learning-based face recognition research. It looks at the MFR pipeline that has been widely used recently and identifies significant advancements that have increased the efficacy of MFR techniques. These advancements are especially focused on overcoming the difficulties caused by barriers created by masks and encouraging the model's applicability to a variety of real-world facial emotions. By refining techniques for creating synthetic datasets and creating novel approaches for gathering authentic face data, researchers and practitioners have achieved tremendous strides. The complicated problems of face detection, face mask recognition, and masked face identification are the focus of these initiatives. At MFR, future research and development should focus on tackling important issues, such as enhancing instruments for producing synthetic datasets and increasing initiatives to gather unique, interesting face data. Improving MFR systems' real-time performance and generalization skills requires overcoming data scarcity. Results will also be enhanced by creating deep learning techniques and investigating novel ideas. Enhancing system accuracy at several stages, including picture preprocessing, feature extraction, face detection and localization, face recovery, identity matching, and authentication, is possible by combining face detection with masked face recognition. By using deep learning models like GAN, CNNs, and SNNs in addition to more conventional machine learning methods like KNN, SVM, and Random Forest, the researchers have achieved impressive results in the extraction and reconstruction of occluded face characteristics from challenging photos. Even with these advancements, a number of restrictions remain that prevent dependable real-time performance in several situations. As these technologies develop and find wider uses, privacy and ethical issues must continue to be at the forefront.

## REFERENCES

[1] M. Fatima, S. A. Ghauri, N. B. Mohammad, H. Adeel, and M. Sarfraz, "Machine Learning for Masked Face Recognition in COVID-19 Pandemic Situation," *Math. Model. Eng. Probl.*, vol. 9, no. 1, pp. 283–289, 2022, doi: 10.18280/mmep.090135.

[2] D. Sai Rohit and A. Professor, "Detection of Masked Faces Using Opencv and Tensorflow," *Int. Res. J. Mod. Eng. Technol. Sci. www.irjmets.com @International Res. J. Mod. Eng.*, no. 07, pp. 2582–5208, 2987, [Online]. Available: www.irjmets.com

[3] Z. Wang and Y. Xu, "Studies advanced in face recognition technology based on deep learning," vol. 0, pp. 121–132, 2023, doi: 10.54254/2755-2721/31/20230133.

[4] H. J. Al-bamarni, N. H. Barnouti, and S. S. M. AL-Dabbagh, "Real-Time Face Detection and Recognition Using Principal Component Analysis (PCA) – Back Propagation Neural Network (BPNN) and Radial Basis Function (RBF)," *J. Theor. Appl. Inf. Technol.*, vol. 91, no. 1, pp. 28–34, 2016.

[5] M. Slavkovic and D. Jevtic, "Face recognition using eigenface approach," *Serbian J. Electr. Eng.*, vol. 9, no. 1, pp. 121–130, 2012, doi: 10.2298/sjee1201121s.

[6] M. Omar, M. Rashedul, and M. Touhid, "Advanced Masked Face Recognition using Robust and Light Weight Deep Learning Model," *Int. J. Comput. Appl.*, vol. 186, no. 2, pp. 42–51, 2024, doi: 10.5120/ijca2024923351.

[7] M. Mahmoud and H. S. Kang, "GANMasker: A Two-Stage Generative Adversarial Network for High-Quality Face Mask Removal," *Sensors*, vol. 23, no. 16, pp. 1–22, 2023, doi: 10.3390/s23167094.

[8] M. Mahmoud, M. S. Kasem, and H. Kang, "applied sciences A Comprehensive Survey of Masked Faces : Recognition , Detection , and Unmasking," 2024.

[9] M. Eman, T. M. Mahmoud, and T. Abd-el-hafeez, "A Novel Hybrid Approach to Masked Face Recognition Using Robust PCA and GOA Optimizer," vol. 13, no. 3, pp. 25–35, 2023, doi: 10.21608/SJDFS.2023.222524.1117.

[10] A. Soni, "Automatic Motorcyclist Helmet Rule Violation Detection using Tensorflow & Keras in OpenCV," no. February, 2020, doi: 10.1109/SCEECS48394.2020.55.

[11] L. Ouannes, A. Ben Khalifa, N. Essoukri, and B. Amara, "Siamese Network for Face Recognition in Degraded Conditions," 2022.

[12] L. Song, D. Gong, Z. Li, C. Liu, and W. Liu, "Occlusion Robust Face Recognition Based on Mask Learning".

[13] R. Alturki, M. Alharbi, and F. Alanzi, "Deep learning techniques for detecting and recognizing face masks : A survey," no. March 2020, 2021.

[14] M. M. Taye, "Understanding of Machine Learning with Deep Learning :," *Comput. MDPI*, vol. 12,

no. 91, pp. 1–26, 2023.
[15] C. Geng and X. Jiang, “Cong Geng and Xudong Jiang School of Electrical and Electronic Engineering Nanyang Technological University , Singapore 639798,” *Rev. Lit. Arts Am.*, pp. 3313–3316, 2009.
[16] C. Liu and H. Wechsler, “Gabor feature based classification using the enhanced Fisher linear discriminant model for face recognition,” *IEEE Trans. Image Process.*, vol. 11, no. 4, pp. 467–476, 2002, doi: 10.1109/TIP.2002.999679.
[17] A. F. Zhilali’l, M. Nasrun, and C. Setianingsih, “Face Recognition Using Local Binary Pattern (LBP) and Local Enhancement (LE) Methods At Night Period,” vol. 2, no. IcoIESE 2018, pp. 103–108, 2019, doi: 10.2991/icoiese-18.2019.19.
[18] Z. Cao, Q. Yin, X. Tang, and J. Sun, “Face recognition with learning-based descriptor,” *Proc. IEEE Comput. Soc. Conf. Comput. Vis. Pattern Recognit.*, no. February, pp. 2707–2714, 2010, doi: 10.1109/CVPR.2010.5539992.
[19] H. Ugail, “Deep Face Recognition,” *Deep Learn. Vis. Comput.*, no. Section 3, pp. 31–47, 2022, doi: 10.1201/9781003091356-4.
[20] 王欣雅, “Masked Face Recognition Based on Deep Learning,” *Comput. Sci. Appl.*, vol. 13, no. 08, pp. 1576–1587, 2023, doi: 10.12677/csa.2023.138156.
[21] A. Alzu, F. Albalas, T. Al-hadhrami, L. B. Younis, and A. Bashayreh, “Masked Face Recognition Using Deep Learning : A Review,” 2021.
[22] W. Rawat and Z. Wang, “Deep Convolutional Neural Networks for Image Classification : A Comprehensive Review Deep Convolutional Neural Networks for Image Classification : A Comprehensive Review,” no. October, 2017, doi: 10.1162/NECO.
[23] L. Omotosho, I. Kazeem, J. Oyeniyi, and O. A. Oyeniran, “A REAL TIME FACE RECOGNITION SYSTEM USING ALEXNET DEEP CONVOLUTIONAL NETWORK TRANSFER LEARNING MODEL A REAL TIME FACE RECOGNITION SYSTEM USING ALEXNET DEEP CONVOLUTIONAL NETWORK TRANSFER LEARNING,” no. September, 2021, doi: 10.29081/jesr.v27i2.277.
[24] J. Sen, B. Sarkar, M. A. Hena, and H. Rahman, “Face Recognition Using Deep Convolutional Network and One-shot Learning,” vol. 7, no. 4, pp. 23–29, 2020.
[25] C. Vimal, “Face and Face-mask Detection System using VGG-16 Architecture based on Convolutional Neural Network,” vol. 183, no. 50, pp. 16–21, 2022.
[26] Y. Pratama, L. M. Ginting, E. Hannisa, and L. Nainggolan, “Face recognition for presence system by using residual networks-50 architecture,” vol. 11, no. 6, pp. 5488–5496, 2021, doi: 10.11591/ijece.v11i6.pp5488-5496.
[27] R. K. Shukla and A. K. Tiwari, “Masked Face Recognition Using MobileNet V2 with Transfer Learning,” no. December 2019, 2023, doi: 10.32604/csse.2023.027986.
[28] C. Google, “Xception: Deep Learning with Depthwise Separable Convolutions,” 2014.
[29] Z. Song, K. Nguyen, T. Nguyen, C. Cho, and J. Gao, “Camera-Based Security Check for Face Mask Detection using Deep Learning,” *Proc. - IEEE 7th Int. Conf. Big Data Comput. Serv. Appl. BigDataService 2021*, no. August 2022, pp. 96–106, 2021, doi: 10.1109/BigDataService52369.2021.00017.
[30] M. Mahmoud, M. S. Kasem, and H.-S. Kang, “A Comprehensive Survey of Masked Faces: Recognition, Detection, and Unmasking,” 2024, doi: 10.3390/app14198781.
[31] F. Kurniawan, I. N. G. A. Astawa, I. W. B. Sentana, I. M. A. D. S. Atmaja, and A. P. Wibawa, “Facemask Detection using the YOLO-v5 Algorithm: Assessing Dataset Variation and Resolutions,” *Regist. J. Ilm. Teknol. Sist. Inf.*, vol. 9, no. 2, pp. 95–102, 2023, doi: 10.26594/register.v9i2.3249.
[32] O. M. Assim and A. F. Mahmood, “Epileptic detection based on deep learning: A review,” *Iraqi J. Electr. Electron. Eng.*, vol. 20, no. 2, pp. 115–126, 2024, doi: 10.37917/ijeee.20.2.10.
[33] N. M. Farhan and B. Setiaji, “Indonesian Journal of Computer Science,” *Indones. J. Comput. Sci.*, vol. 12, no. 2, pp. 284–301, 2023, [Online]. Available: http://ijcs.stmikindonesia.ac.id/ijcs/index.php/ijcs/article/view/3135
[34] P. Liu, S. Han, Z. Meng, and Y. Tong, “Facial expression recognition via a boosted deep belief network,” *Proc. IEEE Comput. Soc. Conf. Comput. Vis. Pattern Recognit.*, pp. 1805–1812, 2014, doi: 10.1109/CVPR.2014.233.
[35] A. P. Kale, R. M. Wahul, A. D. Patange, R. Soman, and W. Ostachowicz, “Development of Deep Belief Network for Tool Faults Recognition,” *Sensors*, vol. 23, no. 4, 2023, doi: 10.3390/s23041872.
[36] N. Ruiz, B. J. Theobald, A. Ranjan, A. H. Abdelaziz, and N. Apostoloff, “MorphGAN: One-Shot Face Synthesis GAN for Detecting Recognition Bias,” *32nd Br. Mach. Vis. Conf. BMVC 2021*, 2021.
[37] Y. Li, L. Song, X. Wu, R. He, and T. Tan, “Anti-makeUp: Learning a bi-level adversarial network

for makeup-invariant face verification," *32nd AAAI Conf. Artif. Intell. AAAI 2018*, pp. 7057–7064, 2018, doi: 10.1609/aaai.v32i1.12294.
[38] D. Yadav, N. Kohli, M. Vatsa, R. Singh, and A. Noore, "Age gap reducer-GaN for recognizing age-separated faces," *Proc. - Int. Conf. Pattern Recognit.*, pp. 10090–10097, 2020, doi: 10.1109/ICPR48806.2021.9412078.
[39] L. Tran, X. Yin, and X. Liu, "Disentangled representation learning GAN for pose-invariant face recognition," *Proc. - 30th IEEE Conf. Comput. Vis. Pattern Recognition, CVPR 2017*, vol. 2017-Janua, pp. 1283–1292, 2017, doi: 10.1109/CVPR.2017.141.
[40] L. Xu, "Face Manipulation with Generative Adversarial Network," *J. Phys. Conf. Ser.*, vol. 1848, no. 1, 2021, doi: 10.1088/1742-6596/1848/1/012081.
[41] H. Yang, D. Huang, Y. Wang, and A. K. Jain, "Learning Face Age Progression: A Pyramid Architecture of GANs," *Proc. IEEE Comput. Soc. Conf. Comput. Vis. Pattern Recognit.*, pp. 31–39, 2018, doi: 10.1109/CVPR.2018.00011.
[42] M. Pudyel and M. Atay, "An Exploratory Study of Masked Face Recognition with Machine Learning Algorithms," *Conf. Proc. - IEEE SOUTHEASTCON*, vol. 2023-April, pp. 877–882, 2023, doi: 10.1109/SoutheastCon51012.2023.10115205.
[43] K. He, X. Zhang, S. Ren, and J. Sun, "Deep residual learning for image recognition," *Proc. IEEE Comput. Soc. Conf. Comput. Vis. Pattern Recognit.*, vol. 2016-Decem, pp. 770–778, 2016, doi: 10.1109/CVPR.2016.90.
[44] C. Li, S. Ge, D. Zhang, and J. Li, "Look through Masks: Towards Masked Face Recognition with De-Occlusion Distillation," *MM 2020 - Proc. 28th ACM Int. Conf. Multimed.*, pp. 3016–3024, 2020, doi: 10.1145/3394171.3413960.
[45] I. Goodfellow *et al.*, "Generative adversarial networks," *Commun. ACM*, vol. 63, no. 11, pp. 139–144, 2020, doi: 10.1145/3422622.
[46] Y. Hu, Y. Xu, H. Zhuang, Z. Weng, and Z. Lin, "Machine Learning Techniques and Systems for Mask-Face Detection—Survey and a New OOD-Mask Approach," *Appl. Sci.*, vol. 12, no. 18, 2022, doi: 10.3390/app12189171.
[47] W. Wang, Z. Zhao, H. Zhang, Z. Wang, and F. Su, "MaskOut: A Data Augmentation Method for Masked Face Recognition," *Proc. IEEE Int. Conf. Comput. Vis.*, vol. 2021-Octob, pp. 1450–1455, 2021, doi: 10.1109/ICCVW54120.2021.00167.
[48] A. Sharma, R. Gautam, and J. Singh, "Deep learning for face mask detection: a survey," *Multimed. Tools Appl.*, vol. 82, no. 22, pp. 34321–34361, 2023, doi: 10.1007/s11042-023-14686-6.
[49] M. Shatnawi, N. Almenhali, M. Alhammadi, and K. Alhanaee, "Deep Learning Approach for Masked Face Identification," *Int. J. Adv. Comput. Sci. Appl.*, vol. 13, no. 6, pp. 296–305, 2022, doi: 10.14569/IJACSA.2022.0130637.
[50] A. Nowrin, S. Afroz, M. S. Rahman, I. Mahmud, and Y. Z. Cho, "Comprehensive Review on Facemask Detection Techniques in the Context of Covid-19," *IEEE Access*, vol. 9, pp. 106839–106864, 2021, doi: 10.1109/ACCESS.2021.3100070.
[51] E. Randellini, L. Rigutini, and C. Saccà, "Data Augmentation Techniques and Transfer Learning Approaches Applied to Facial Expressions Recognition Systems," *Int. J. Artif. Intell. Appl.*, vol. 13, no. 1, pp. 55–72, 2022, doi: 10.5121/ijaia.2022.13104.
[52] K. N. N. Classifier, M. Eman, T. M. Mahmoud, and M. M. Ibrahim, "Using Pretrained Mask Detection and Segmentation , Robust," 2023.
[53] A. K. Sharadhi, V. Gururaj, S. P. Shankar, M. S. Supriya, and N. S. Chogule, "Face mask recogniser using image processing and computer vision approach," *Glob. Transitions Proc.*, vol. 3, no. 1, pp. 67–73, 2022, doi: 10.1016/j.gltp.2022.04.016.
[54] B. Qin and D. Li, "Identifying facemask-wearing condition using image super-resolution with classification network to prevent COVID-19," *Sensors (Switzerland)*, vol. 20, no. 18, pp. 1–23, 2020, doi: 10.3390/s20185236.
[55] D. C. L. Ngo, A. B. J. Teoh, and A. Goh, "Biometric hash: High-confidence face recognition," *IEEE Trans. Circuits Syst. Video Technol.*, vol. 16, no. 6, pp. 771–775, 2006, doi: 10.1109/TCSVT.2006.873780.
[56] B. Olushola, "Overview of Biometric and Facial Recognition Techniques Omoyiola, Bayo Olushola," *IOSR J. Comput. Eng.*, vol. 20, no. 4, pp. 1–05, 2018, doi: 10.9790/0661-2004010105.
[57] U. Bakshi and R. Singhal, "A Survey on Face Detection Methods and Feature Extraction Techniques of Face Recognition," *Int. J. Emerg. Trends Technol. Comput. Sci.*, vol. 3, no. 3, pp. 233–237, 2014.
[58] K. Y. Houssamddine Hamdi, "MASKED FACE RECOGNITION BASED ON FaceNet PRE-TRAINED MODEL," *Kamil Yurtkan*, no. January, 2022.
[59] B. Huang *et al.*, "Masked Face Recognition Datasets and Validation," *Proc. IEEE Int. Conf. Comput.*

*Vis.*, vol. 2021-Octob, pp. 1487–1491, 2021, doi: 10.1109/ICCVW54120.2021.00172.

[60] Z. Wang, B. Huang, G. Wang, P. Yi, and K. Jiang, "Masked Face Recognition Dataset and Application," *IEEE Trans. Biometrics, Behav. Identity Sci.*, vol. 5, no. 2, pp. 298–304, 2023, doi: 10.1109/TBIOM.2023.3242085.

[61] D. Yi, Z. Lei, S. Liao, and S. Z. Li, "Learning Face Representation from Scratch," 2014, [Online]. Available: http://arxiv.org/abs/1411.7923

[62] H. Deng, Z. Feng, G. Qian, X. Lv, H. Li, and G. Li, "MFCosface: a masked-face recognition algorithm based on large margin cosine loss," *Appl. Sci.*, vol. 11, no. 16, 2021, doi: 10.3390/app11167310.

[63] X. Jiang, T. Gao, Z. Zhu, and Y. Zhao, "Real-time face mask detection method based on yolov3," *Electron.*, vol. 10, no. 7, pp. 1–17, 2021, doi: 10.3390/electronics10070837.

[64] A. Kantarcı, F. Ofli, M. Imran, and H. K. Ekenel, "Bias-Aware Face Mask Detection Dataset," 2022, [Online]. Available: http://arxiv.org/abs/2211.01207

[65] S. Mishra, P. Majumdar, R. Singh, and M. Vatsa, "Indian Masked Faces in the Wild Dataset," *Proc. - Int. Conf. Image Process. ICIP*, vol. 2021-Septe, pp. 884–888, 2021, doi: 10.1109/ICIP42928.2021.9506447.

[66] R. Golwalkar and N. Mehendale, "Masked Face Recognition Using Deep Metric Learning and FaceMaskNet-21," *SSRN Electron. J.*, 2020, doi: 10.2139/ssrn.3731223.

[67] M. Knoche, S. Hormann, and G. Rigoll, "Cross-Quality LFW: A Database for Analyzing Cross-Resolution Image Face Recognition in Unconstrained Environments," *Proc. - 2021 16th IEEE Int. Conf. Autom. Face Gesture Recognition, FG 2021*, pp. 1–9, 2021, doi: 10.1109/FG52635.2021.9666960.

[68] Q. Cao, L. Shen, W. Xie, O. M. Parkhi, and A. Zisserman, "VGGFace2: A dataset for recognising faces across pose and age," *Proc. - 13th IEEE Int. Conf. Autom. Face Gesture Recognition, FG 2018*, pp. 67–74, 2018, doi: 10.1109/FG.2018.00020.

[69] Y. Xin *et al.*, "Machine Learning and Deep Learning Methods for Cybersecurity," *IEEE Access*, vol. 6, no. May 2021, pp. 35365–35381, 2018, doi: 10.1109/ACCESS.2018.2836950.

[70] M. Kubat, *An Introduction to Machine Learning*. 2017. doi: 10.1007/978-3-319-63913-0.

[71] A. Tharwat, "Classification assessment methods," *Appl. Comput. Informatics*, vol. 17, no. 1, pp. 168–192, 2018, doi: 10.1016/j.aci.2018.08.003.

[72] S. Shaik, C. Konda, and M. Tech, "a Noval Approach To Iris Recognition Based on Feature Level Fusion Using Classification Techniques," *Int. J. Sci. Dev. Res.*, vol. 2, no. 7, pp. 287–292, 2017, [Online]. Available: www.ijsdr.org

[73] T. F. Monaghan *et al.*, "Foundational statistical principles in medical research : Predictive value," *Medicina (B. Aires).*, vol. 57, no. 5, p. 503, 2021.

[74] J. S. Akosa, "Predictive Accuracy: A Misleading Performance Measure for Highly Imbalanced Data," *SAS Glob. Forum*, vol. 942, pp. 1–12, 2017, [Online]. Available: https://support.sas.com/resources/papers/proceedings17/0942-2017.pdf

[75] I. M. De Diego, A. R. Redondo, R. R. Fernández, J. Navarro, and J. M. Moguerza, "General Performance Score for classification problems," *Appl. Intell.*, vol. 52, no. 10, pp. 12049–12063, 2022, doi: 10.1007/s10489-021-03041-7.

[76] R. Jahangir, Y. W. Teh, H. F. Nweke, G. Mujtaba, M. A. Al-Garadi, and I. Ali, "Speaker identification through artificial intelligence techniques: A comprehensive review and research challenges," *Expert Syst. Appl.*, vol. 171, 2021, doi: 10.1016/j.eswa.2021.114591.

[77] M. H. Murray and J. D. Blume, "False Discovery Rate Computation: Illustrations and Modifications," pp. 1–18, 2020, [Online]. Available: http://arxiv.org/abs/2010.04680

[78] J. Terven, D. M. Cordova-Esparza, A. Ramirez-Pedraza, E. A. Chavez-Urbiola, and J. A. Romero-Gonzalez, "Loss Functions and Metrics in Deep Learning," 2023, [Online]. Available: http://arxiv.org/abs/2307.02694